\documentclass{article}

\usepackage{arxiv}

\usepackage[utf8]{inputenc} 
\usepackage[T1]{fontenc}    
\usepackage{hyperref}       
\usepackage{url}            
\usepackage{booktabs}       
\usepackage{amsfonts}       
\usepackage{nicefrac}       
\usepackage{microtype}      
\usepackage{lipsum}
\usepackage{graphicx}
\usepackage{color}
\usepackage{caption}

\title{Independent Vector Analysis for Data Fusion \\Prior to Molecular Property Prediction \\with Machine Learning}

\author{
	Zois Boukouvalas \\
	University of Maryland, College Park\\
	Dept. of Mechanical Engineering\\
	College Park, MD 20742 \\
	\texttt{zoisb@umd.edu} \\
	\And
	Daniel C. Elton\\
	University of Maryland, College Park\\
	Dept. of Mechanical Engineering\\
	College Park, MD 20742 \\
	\texttt{delton@umd.edu} \\
	\AND
	Peter W. Chung\\
	University of Maryland, College Park\\
	Dept. of Mechanical Engineering\\
	College Park, MD 20742 \\
	\texttt{pchung15@umd.edu} \\
	\And
	Mark D. Fuge\\
	University of Maryland, College Park\\
	Dept. of Mechanical Engineering\\
	College Park, MD 20742 \\
	\texttt{fuge@umd.edu} 
}

\begin{document}
\maketitle

\begin{abstract}
Due to its high computational speed and accuracy compared to ab-initio quantum chemistry and forcefield modeling, the prediction of molecular properties using machine learning has received great attention in the fields of materials design and drug discovery. A main ingredient required for machine learning is a training dataset consisting of molecular features\textemdash for example fingerprint bits, chemical descriptors, etc. that adequately characterize the corresponding molecules. However, choosing features for any application is highly non-trivial. No ``universal'' method for feature selection exists. In this work, we propose a data fusion framework that uses Independent Vector Analysis to exploit underlying complementary information contained in different molecular featurization methods, bringing us a step closer to automated feature generation. Our approach takes an arbitrary number of individual feature vectors and automatically generates a single, compact (low dimensional) set of molecular features that can be used to enhance the prediction performance of regression models. At the same time our methodology retains the possibility of interpreting the generated features to discover relationships between molecular structures and properties. We demonstrate this on the QM7b dataset for the prediction of several properties such as atomization energy, polarizability, frontier orbital eigenvalues, ionization potential, electron affinity, and excitation energies. In addition, we show how our method helps improve the prediction of experimental binding affinities for a set of human BACE-1 inhibitors. To encourage more widespread use of IVA we have developed the PyIVA Python package, an open source code which is available for download on Github.\end{abstract}

\section{Introduction}
	Machine learning (ML) has recently been used for the prediction of molecular properties and studies have shown that it can provide accurate and computationally efficient solutions for this task \cite{rupp2012fast,hansen2015machine,Elton2018scirep, BarnesIDS2018, 1367-2630-15-9-095003, barker2016localized, faber2017machine}. The prediction ability of a ML model highly depends on the proper selection of a training dataset, via feature vectors, that can fully capture certain characteristics of a given set of molecules. A common way to represent a molecule is through a string representation called the Simplified Molecular-Input Line-Entry System (SMILES) string \cite{weininger1988smiles}. Although working directly with SMILES strings has shown to be effective in some ML tasks \cite{Sanchez-Lengeling2018:sciencemag}, most of the ML methods require vector or matrix variables, rather than strings.
	Basic classes of featurization methods that have been widely used in the literature include cheminformatic descriptors, molecular fingerprints, and custom graph convolution based fingerprints, where each featurization method provides different -- though not necessarily unique -- information about a molecule. 
	
	
	Thus it is important to find answers to the question "how can disparate datasets, each associated with unique featurization methods, be integrated?"  The question is motivated by the desire to create automated approaches where data generation techniques are integrable with ML models. \emph{Data fusion} methods \cite{hall1997introduction, castanedo2013review} may serve this purpose since they enable simultaneous study of multiple datasets by, for instance, exploiting alignments of data fragments where there is a common underlying latent space.
	A naive approach could be to simply concatenate data that has been generated by different featurization methods. However, this typically leads to a curse of dimensionality that could affect the performance of a ML algorithm and make it impossible to discover the features of greatest importance. Therefore, selecting a model that generates a set of molecular feature vectors by effectively exploiting complementary information among multiple datasets is an important issue. 
	
	
	Blind source separation (BSS) techniques enable the joint analysis of datasets and extraction of summary factors with few assumptions placed on the data. This is generally achieved through the use of a generative model. One of the most widely used BSS techniques is independent component analysis (ICA) \cite{comon2010handbook,AdaliDiversity}. The popularity is likely because by only requiring statistical independence of the latent sources it can uniquely identify the true latent sources subject to only scaling and permutation ambiguities. However, ICA can only decompose a single dataset. In many applications multiple sets of data are gathered about the same phenomenon.  These sets can often be disjoint and merely misaligned fragments of a shared underlying latent space. Multiple datasets could therefore share dependencies. This motivates the application of methods that can jointly factorize multiple datasets, like independent vector analysis (IVA) \cite{4032777}. IVA is a recent generalization of ICA to multiple datasets that can achieve improved performance over performing ICA on each dataset separately. 
	
	In this paper, we propose a novel data fusion framework that uses IVA to exploit underlying complementary information contained across multiple datasets.  Namely, we show that information that has been generated by different molecular featurization methods can be fused to improve learning of the chemical relationships. 
	Note here that achieving perfect regression error is not the the goal of this work.  Instead, our goal is to determine how to generate feature vectors by combining datasets from multiple featurization methods to improve the learned response of the data-fused model over the performance of the individual feature vectors treated separately. 
	Particularly noteworthy is that the proposed approach is parameter free, computationally attractive when compared with existing methods \cite{1367-2630-15-9-095003}, easily interpretable due to the simplicity of the generative model, and does not require a large amount of training samples in order to achieve a desirable regression error.
	
	The remainder of this paper is organized as follows. In Section 2, we provide a brief background on the IVA method and the regression procedure. Section 3 provides the results of the regression procedure and associated discussions. The conclusions and future research directions are presented in Section 4. 
	

\section{Materials and Methods}
	The BSS model is formulated as follows. Let ${\bf X}\in \mathbb{R}^{d\times N}$ be the observation matrix where $d$ denotes the dimension of the feature vector and $N$ denotes the total number of molecules. The noiseless BSS generative model is given by
	\begin{equation}\label{OriginalBSS}
	{\bf X} = {\bf A}{\bf S},
	\end{equation}
	where ${\bf A}\in \mathbb{R}^{d\times P}$ is the mixing matrix, and ${\bf S}\in \mathbb{R}^{P\times N}$ is the matrix that contains the sources that need to be estimated and will be used as the new feature vector for the ML task. One of the most widely used methods for solving the BSS problem (\ref{OriginalBSS}) is ICA and its basic assumption is that the source signals are statistically independent \cite{bell1997independent, hyvarinen2000independent, ristaniemi1999performance, back1997first,lee2007fast}. 
	
	By rewriting (\ref{OriginalBSS}) using random vector notation, we have
	\begin{equation}\label{BSS}
	{\bf x}(n) = {\bf A}{\bf s}(n),~~~ n=1,\dots,N, \nonumber
	\end{equation}
	where $n$ is the sample index denoting the $n$th molecule, ${\bf s}(n)\in \mathbb{R}^{P}$ are the unknown sources that need to be estimated, and ${\bf x}(n)\in \mathbb{R}^{d}$ are the mixtures. Our interest is in dealing with overdetermined problems where $d>P$. This can be reduced to the case where $d = P$ using a dimensionality reduction technique like principal component analysis (PCA). For the purpose of this work and for the rest of this paper we assume that the samples of each unknown source are independent and identically distributed and therefore, to simplify the notation we drop the sample index $n$. 
	
	Although ICA has been shown to be very useful in many applications, it decomposes a single dataset at a time. For practical applications that involve more than one dataset one could perform ICA separately on each dataset and align the subsequent results. However, this approach could be considered suboptimal since performing ICA individually on each dataset will ignore any dependencies that exist among them.
	
	\subsection{Independent Vector Analysis}
	It is common for datasets that have been generated from different featurization methods to have some inherent dependence among them. IVA generalizes the ICA problem by allowing for full exploitation of this dependence leading to improved performance beyond what is achievable by applying ICA separately to each dataset. Additionally, IVA automatically aligns dependent sources across the datasets, thus bypassing the need for a second permutation correction algorithm.
	
	IVA is similar to ICA, except that now we have $K$ datasets, ${{\bf x}}^{[k]}$, $k=1,...,K$ where each dataset is a linear mixture of $N$ statistically independent sources. 
	Under the assumption that PCA preprocessing results in $d=P$, the noiseless IVA model is given by
	\begin{equation}\nonumber
	{{\bf x}}^{[k]} = {\bf A}^{[k]} {\bf s}^{[k]},~~~ k=1,...,K,
	\end{equation}
	where ${\bf A}^{[k]}\in \mathbb{R}^{P\times P},\ k = 1,...,K$ are invertible mixing matrices and ${\bf s}^{[k]} = [ s_1^{[k]},...,s_P^{[k]}]^{\top}$ is the vector of latent sources for the $k$th dataset. In the IVA model, the components within each ${\bf s}^{[k]}$ are assumed to be independent, while at the same time, dependence across corresponding components of ${\bf s}^{[k]}$ is allowed. To mathematically formulate the dependence across components that IVA can take into account, we define the source component vector (SCV) by vertically concatenating the $p$th source from each of the $K$ dataset as
	\begin{equation}
	{\bf s}_p = [s_p^{[1]},...,s_p^{[K]}]^{\top},
	\end{equation}
	where ${\bf s}_p$ is a $K$-dimensional random vector. The goal in IVA is to estimate $K$ demixing matrices to yield source estimates ${\bf y}^{[k]} = {\bf W}^{[k]} {\bf x}^{[k]}$, such that each SCV is maximally independent of all other SCVs. It should be mentioned that dependence across datasets is not a necessary condition for IVA to work. In the case where this type of statistical property does not exist, IVA reduces to individual ICAs on each dataset.
	
	The IVA cost function can be defined in a similar manner as ICA. However, the optimization parameter is not just a single demixing matrix ${\bf W}$ as in the ICA case, but a set of demixing matrices ${\bf W}^{[1]},\ldots,{\bf W}^{[K]}$, which can be collected into a three dimensional array ${\cal W}\in \mathbb{R}^{P\times P\times K}$. The IVA objective function is given by 
	\begin{equation}\label{IVAcostfunctionDiversity}
	J_{IVA}({\cal W}) = \sum_{p=1}^P H({\bf y}_p) - \sum_{k=1}^K \log \left|\det \left( {\bf W}^{[k]}\right)\right| - H ({\bf x}^{[1]},...,{\bf x}^{[K]}),
	\end{equation}
	where $H({\bf y}_p)$ denotes the differential entropy of the estimated $p$th SCV and the term $H({\bf x}^{[1]},...,{\bf x}^{[K]})$ is a constant parameter where it can be treated as a constant for optimization purposes. By definition, the term $H({\bf y}_p)$ equals $\displaystyle\sum_{k=1}^K H(y_p^k) - I({\bf y}_p)$, where $I({\bf y}_p)$ denotes the mutual information within the $p$th SCV. Therefore, it can be observed that minimization with respect to each demixing matrix ${\bf W}^{[k]}$ of (\ref{IVAcostfunctionDiversity}) automatically increases the mutual information within the components of a SCV, revealing how IVA exploits this type of statistical property. It can be seen that without the mutual information term, the objective function (\ref{IVAcostfunctionDiversity}) is equivalent to performing independent ICA separately on each dataset. For more information about the derivation of the IVA cost function and the optimization scheme, we refer the reader to \cite{ZoisThesis}.
	
	From (\ref{IVAcostfunctionDiversity}) it can be observed that the probability density function or its approximation for each estimated SCV plays an important role on the estimation performance of the demixing matrices. One of the first IVA algorithms was formulated for solving the convolutive ICA problem in the frequency domain using multiple frequency bins \cite{lee2007fast}. This led to the development of IVA-Laplacian (IVA-L) \cite{4032777}, an algorithm that takes higher-order statistics (HOS) into account and assumes a Laplacian distribution for the underlying source component vectors. Due to its simplicity and computational efficiency, for the purpose of our work we use IVA-L. 
	
	\subsection{Feature extraction and regression procedure}\label{RegressionFramework}
	The regression process consists of three stages. In the first stage, $N$ molecules are randomly sub sampled in order to generate training and testing datasets, ${\bf X}^k_{\rm Train}$ and ${\bf X}^k_{\rm Test}$ respectively, for each $k=1,\dots,K$. 
	
	In the second stage, the mean from each dataset is removed and PCA is applied to each ${\bf X}^k_{\rm Train}$ using an order $P$, implying that the signal subspace contains the components that have higher variance. Then for each $k\in\{1,\dots,K\}$, we generate $\hat{\bf X}^k_{\rm Train}\in \mathbb{R}^{P\times N_{\rm Train}}$ and by vertically concatenating each $\hat{\bf X}^k_{\rm Train}$ we form a three dimensional array ${\hat{\bf X}}_{\rm Train}\in\mathbb{R}^{P\times N\times K}$. IVA-L is performed on ${\hat{\bf X}}_{\rm Train}$ resulting in a set of demixing matrices, $\{ {\bf W}^1,\dots,{\bf W}^K \}$, where each ${\bf W}^k\in\mathbb{R}^{P\times P}$. Using the demixing matrices we generate ${\bf Y}^k_{\rm Train} = {\bf W}^k({\hat{\bf X}}^{k}_{\rm Train})^{\top}$ for each $k=1\dots,K$ and the training dataset ${\bf Y}_{\rm Train}$ is formed by vertically concatenating the SCVs, ${\bf y}_1,\dots,{\bf y}_P$. 
	
	The testing dataset is generated by removing the mean from each testing dataset and by using the PCA training transformation from the training phase, we generate ${\hat{\bf X}}^k_{\rm Train},\ k=1\dots,K$. Each of the demixing matrices from the training phase is used to create testing datasets ${\bf Y}^k_{\rm Test}\in \mathbb{R}^{P\times N_{\rm Test}}$ by ${\bf Y}^k_{\rm Test} = {\bf W}^k({\hat{\bf X}}^{k}_{\rm Test})^{\top}$ for each $k=1\dots,K$. Finally, the testing dataset is formed by vertically concatenating the testing SCVs. 
	
	In the third stage, we train the regression model using $({\bf Y}_{\rm Train})^{\top}$. Here, the specific form of the regression function is unimportant. But to demonstrate a concrete example, we use kernel ridge regression (KRR) with a Gaussian kernel, which follows from previous work \cite{Elton2018scirep} and \cite{rupp2012fast}. Once the regression model has been trained we evaluate its performance by using the unseen data $({\bf Y}_{\rm Test})^{\top}$. For all of the experiments, hyperparameter optimization and model training and testing is done using a nested cross validation scheme. In the inner loop, the length scale parameter of the Gaussian kernel and the regularization parameter are optimized using grid search selection using $80\%$ of the data to train the model and another $10\%$ for validation. The remaining $10\%$ is held out as a test set to estimate performance after hyperparameter optimization. The outer loop is done five times, corresponding to five folds. This entire process was repeated 30 times (with shuffling before each iteration) to generate well-converged statistics. 
	
	\subsection{Data Sources and Methods}
	The scope of the present work is limited to the data from two sources: the QM7b dataset \cite{blum,1367-2630-15-9-095003} and the human $\beta$-secretase-1 (BACE-1) inhibitors dataset from the \href{moleculenet.ai}{\textit{moleculenet.ai}} website \cite{Wu2018:513}. 
	
	The QM7b dataset contains 7211 organic molecules made of up to 7 heavy atoms. These atoms include C, N, O, S, and Cl.The only other element present in the molecules is hydrogen. For each molecule several properties have been calculated at different levels of theory, including density functional theory and the many-body GW approach. The properties include the atomization energy, polarizability, HOMO and LUMO eigenvalues, and excitation energy. From the QM7b data, we generate three datasets using three different featurization methods: Coulomb matrices eigenspectra (CME) \cite{rupp2012fast}, sum over bonds (SOB) \cite{Elton2018scirep}, and weight matrices eigenspectra (WE). The Coulomb matrix of a molecule is specified by the 3-dimensional coordinates of the atoms as well as their atomic charges. Since the Coulomb matrix is not invariant under row or column permutations, the eigenspectra are used as a feature vector. The dimensionality of the Coulomb matrix ($d = 23$ for QM7b) is set by the molecule with the largest number of atoms in the dataset. Sum over bonds is defined as a bond count and the feature vectors are generated by first enumerating all of the bond types in the dataset and then counting how many of each bond are present in each molecule ($d = 28$ for QM7b). To compute the weight matrix eigenspectra (WE) featurization, each molecule is treated as a graph, where edges represent bonds and vertices represent atoms. The value of each matrix element represents the bond order, e.g. `1' denotes a single bond, `2' a double bond etc. Similarly to the Coulomb matrix featurization method, for each molecule we define the feature vector to be the eigenspectrum of its weight matrix, and the dimension of the WE feature vector is $d = 23$.
	
	The BACE-1 inhibitor dataset is obtained from the \href{moleculenet.ai}{\textit{moleculenet.ai}} website.\cite{Wu2018:513} 1,513 of the values were taken from an earlier work \cite{Subramanian2016:1936}, which in turn obtained them from over thirty different laboratories. We then created seven  basic datasets using the descriptors contained in the MoleculeNet dataset and descriptors generated with the \textit{mmltoolkit} \cite{mmltoolkit} and \textit{RDKit} \cite{rdkit}. The first dataset ($d=13$) consists of physio-chemical descriptors computed using BIOVIA's Pipeline Pilot package. The second dataset ($d=32$) consists of additional Canvas descriptors.\cite{CanvasWebpage} The third dataset ($d=28$) uses the SOB featurization method. The fourth dataset ($d=88$) consists of counts of each atom type as defined in the electrotopological state index (the ``E-state'' count fingerprint).\cite{Kier1995:1039}. Dataset five ($d=34$) consists of the RDKit molecular surface charge descriptors \cite{rdkit,Ertl2000:3714}. Dataset six ($d=60$) consists of the RDKit functional group counts and dataset seven ($d=28$) consists of an additional custom set of functional group counts obtained using the $\textit{functional\_group\_featurizer}$ in the \textit{mmltoolkit}.  We then use these basic seven sets to create $K=7$ different combinations of sets.  Therefore, the case where $K=2$ consists of 21 different combinations from the seven basic sets, and when $K=3$ there are 35 combinations, etc.  
	

\section{Results and Discussion}
	The QM7b dataset study examines a) regression performance with respect to three fusing methods and the size of the datasets and b) interpretability properties of IVA. The relatively large size of the dataset and the number of chemical properties make QM7b desirable for this part of the study. The BACE-1 inhibitor dataset is used to demonstrate how different featurization methods complement each other and to contrast the effects of different combinations of featurizations on the performance of the regression model. 
	
	\subsubsection{Regression analysis}
	The first three columns in Table (\ref{table1Reference}) show the MAEs for out-of-sample predictions by the KRR model for each type of featurization method. Errors are reported for the 14 properties associated with the QM7b dataset. From (\ref{table1Reference}), we see that SOB provides the best results except for $\alpha$(SCS), where CME is more accurate. Overall, WE performs the worst.  These results are not surprising since WE contains information only about the bonds within a molecule and is the only featurization method that lacks information about the atomic species in each molecule, which is critical to many of the properties. 
	
	The last three columns of Table (\ref{table1Reference}) list the MAEs when different techniques have been used to generate feature vectors for training the regression model. The so-called Regular approach denotes the procedure where the SOB, CME, and WE datasets are vertically concatenated resulting in $N$ feature vectors of dimension $74$. The ICA and IVA approaches share the same feature extraction and regression procedures described in (\ref{RegressionFramework}). The only difference is that after we apply PCA to each ${\bf X}^k_{\rm Train}$, we perform ICA by entropy bound minimization (ICA-EBM) \cite{5499122} on each $\hat{\bf X}^k_{\rm Train}$ training dataset separately. For both ICA and IVA approaches the resulting $N$ feature vectors are of dimension $30$. From Table (\ref{table1Reference}), we observe that IVA leads to an improved MAE for 8 out of 14 properties when compared to the ``Regular'' approach. For all properties we have examined, IVA yields a lower MAE than ICA.  This is likely due to IVA's ability to resolve separate demixing matrices for each of the $K$ data sets, thereby exploiting the complementary information that is shared across them. 
	
	\begin{table}
		\caption{Average MAE for different featurization methods and different approaches to generate feature vectors for 14 properties.}
		\label{table1Reference}
		\centering
		\begin{tabular}{|c | c || c | c | c || c | c | c |}
				\hline
				Property & Units & SOB & CME & WE & Regular & ICA & IVA\\
				\hline\hline
				$E$(PBEO) & Kcal/mol & 6.029 & 9.966 & 44.398 & {\bf 2.662} & 3.187 & 2.989 \\  
				
				$E^*_{{\rm max}}$ & eV & 1.206 & 2.076 & 1.65 & 1.423 &1.478 & {\bf 1.335} \\ 
				
				$I_{{\rm max}}$ & Arbitrary & 0.082 & 0.105 & 0.106 &0.081 & 0.08 & {\bf 0.074} \\
				
				HOMO(ZINDO) & eV & 0.255 & 0.488 & 0.537 & {\bf 0.183} & 0.206 & 0.194\\
				
				LUMO(ZINDO) & eV & 0.181 & 0.57 & 0.529 & 0.124 & 0.111 & {\bf 0.105}\\
				
				$E^*_{1^{st}}$(ZINDO) & ev & 0.266 & 0.693 & 0.714 & 0.176 & 0.183 & {\bf 0.174} \\
				
				IP(ZINDO) & eV & 0.279 & 0.507 & 0.551 & {\bf 0.225} & 0.246 & 0.229\\
				
				EA(ZINDO) & eV & 0.181 & 0.632 & 0.575 & 0.135 & 0.12 & {\bf 0.113}\\
				
				HOMO(PBEO) & eV & 0.232 & 0.344 & 0.41 & {\bf 0.18} & 0.207 & 0.19 \\
				
				LUMO(PBEO) & eV & 0.183 & 0.304 & 0.324 & 0.132 & 0.141 & {\bf 0.128}\\
				
				HOMO(GW) & eV & 0.254 & 0.375 & 0.446 & {\bf 0.195} & 0.222 & 0.205\\
				
				LUMO(GW) & eV & 0.182 & 0.235 & 0.242 & 0.154 & 0.156 & {\bf 0.141}\\
				
				$\alpha$(PBEO) & {\AA}$^3$ & 0.224 & 0.308 & 0.472 & 0.126 & 0.136 & {\bf 0.123}\\
				
				$\alpha$(SCS) & {\AA}$^3$ & 0.259 & 0.202 & 0.434 & {\bf 0.086} & 0.112 & 0.1\\
				\hline
		\end{tabular}
	\end{table}
	
	We also explored the convergence behavior of the regression model as the number of molecules in the training dataset is increased. Plots of MAE as a function of training points $N$ are called learning curves and typically follow a power law MAE$ = C * N^{\alpha}$, where $C$ is a constant and $\alpha$ is the power law exponent which we refer to as the rate of convergence. These results suggest that IVA supports small data problems well.  However, when large data are available, a fusion approach becomes less beneficial due to the smaller convergence rate.
	
\begin{figure}[h]
		\centering
		\includegraphics[width = 0.9\textwidth]{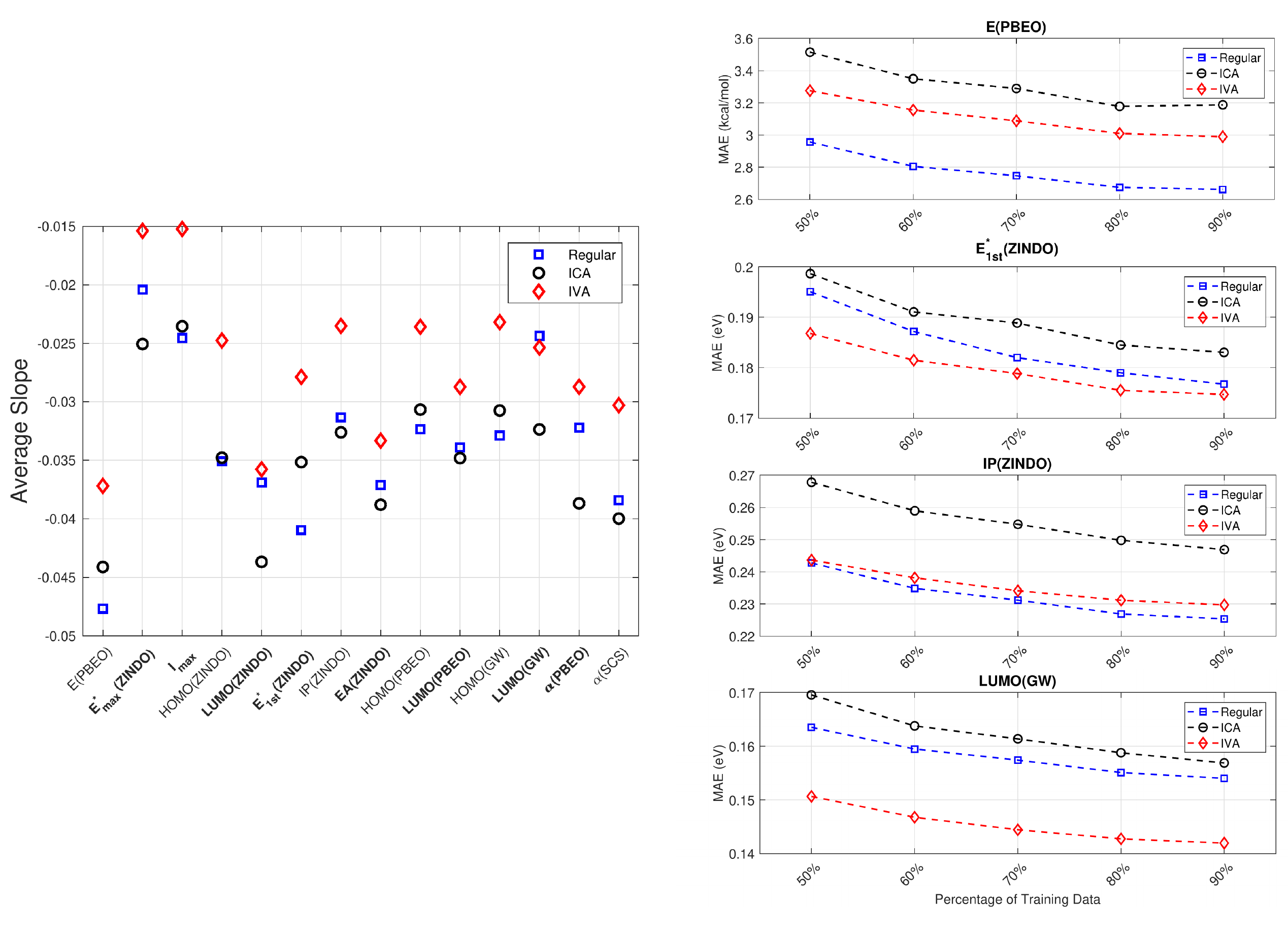}
		\caption{On the left: Slope of the learning curve on a log-log plot. On the right: Representative examples showing how the relative performance of IVA, ICA, and regular approach can be significantly different depending on the target property being predicted.}
		\label{DecreasePlot}
	\end{figure}

	The calculated rates are shown in Figure (\ref{DecreasePlot}) for each approach and for several representative properties. We observe that IVA has a smaller rate of convergence as a function of sample size but lower absolute error through most of the range of sample sizes we examined.  This indicates that the benefits of fusing data are best when the data are small and that a likely cross-over point occurs where the benefits of fusing data are outweighed by the slower convergence. On the left in Figure (\ref{DecreasePlot}) we mark in bold the properties where IVA provides a lower MAE when the full dataset is used. On the right in Figure (\ref{DecreasePlot}), we show the MAE obtained after five-fold cross-validation as a function of different $N$ for four representative target properties. For the most part, absolute errors from IVA are substantially lower. The degree to which the absolute error is smaller depends on the target property and data size.  For instance, the Regular approach gives the lowest MAE for atomization energy (E(PBE0)). For $E^*_{1{\rm st}}$(ZINDO), IVA gives the lowest MAE, but we see that as $N$ increases the gains from using IVA decrease as more data is added indicating that the the curse of dimensionality becomes less of an issue for larger $N$. For IP(ZINDO) the slope of both IVA and regular are approximately the same. Finally, for LUMO(GW) we see a large improvement from IVA, and ICA and the Regular approach tend to achieve the same MAE as $N$ increases.   
	
	\subsubsection{Significance of mixing matrices}
	An important property of IVA is that it can also
	provide chemical or physical interpretations through the estimated mixing matrix. Once the demixing matrices, ${\bf W}^{[k]}$ have been estimated using the approach in Section \ref{RegressionFramework}, we can estimate the mixing matrix by performing back-reconstruction. In order to describe this procedure remember that before IVA is performed, PCA is applied to each ${\bf X}^{k}$ using a presumed order $P$. This provides the data reduction matrix ${\bf F}^{k}$ where each ${\bf F}^{k}$ is formed by the eigenvectors with the first $P$ highest eigenvalues of the corresponding ${\bf X}^{k}$. An estimate of the mixing matrix ${\bf A}^k$ is computed as ${\hat{\bf A}}^{k} = ({\bf F}^{k})^{\dagger}({\bf W}^{k})^{-1}$, where $(\cdot)^{\dagger}$ denotes the pseudo-inverse of a matrix. 
	
\begin{figure}[h]
		\centering
		\includegraphics[width = 0.85\textwidth]{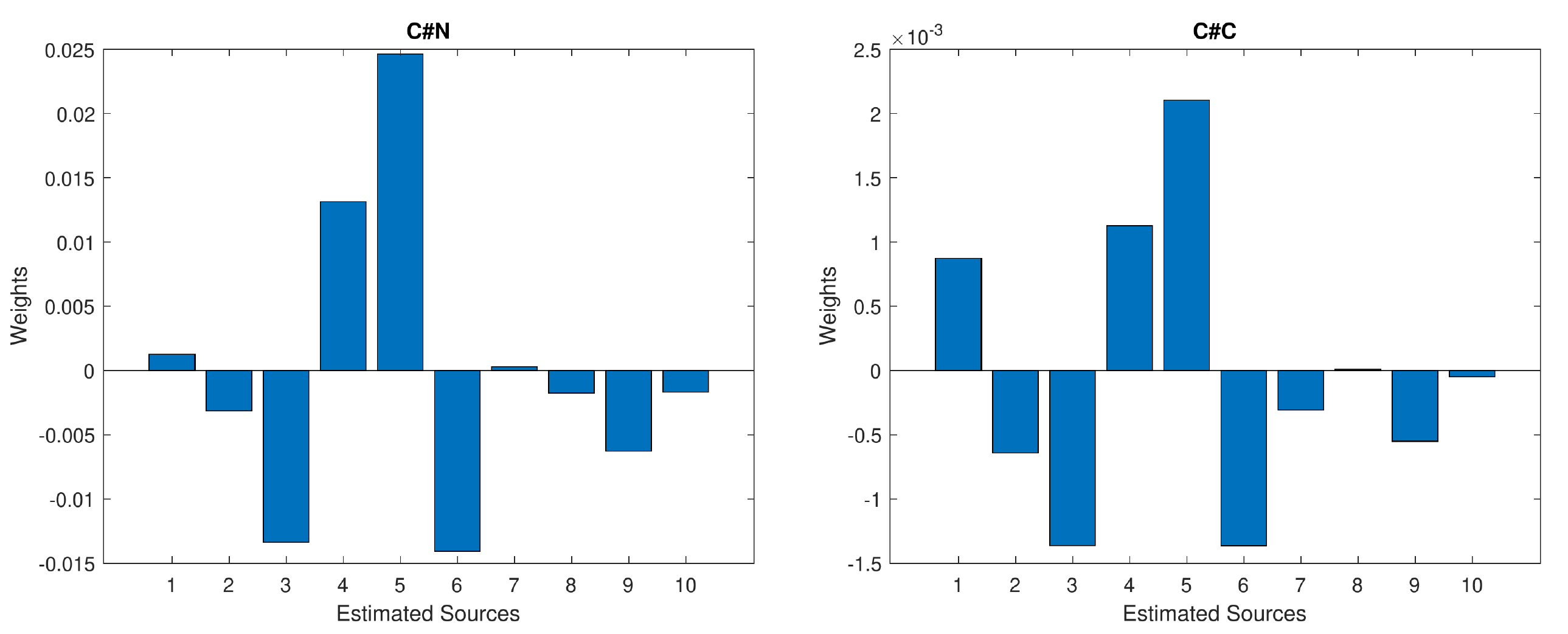}
		\caption{Weights for different features in SOB as a function of the estimated sources. Bars indicate the value of each element in the corresponding row of the estimated mixing matrix.}
		\label{TripleBonds}
	\end{figure}

	Each row of the $k$th estimated demixing matrix therefore represents the weights for the estimated sources of the $k$th dataset. The values of the weights can reveal relationships between certain characteristics of a given set of molecules. The weights in the mixing matrix from the SOB features are shown in Figure (\ref{TripleBonds}).  A common motif appears for the weights associated with the CN and CC triple bond.  Other motifs appear for other molecular features, which we are in the process of analyzing further. This provides early evidence that the IVA generative model encodes information regarding the relationships among different molecular features. 
	
	\subsection{(BACE-1) inhibitors dataset}
	We next evaluate the performance of the IVA approach as a function of the number of datasets that are fused.  The specific targets of the calculation are the experimental binding affinities (IC50 values) for a set of 1,522 human $\beta$-secretase-1 (BACE-1) inhibitors.
	
\begin{figure}[h]
		\centering
		\includegraphics[width = 0.45\textwidth]{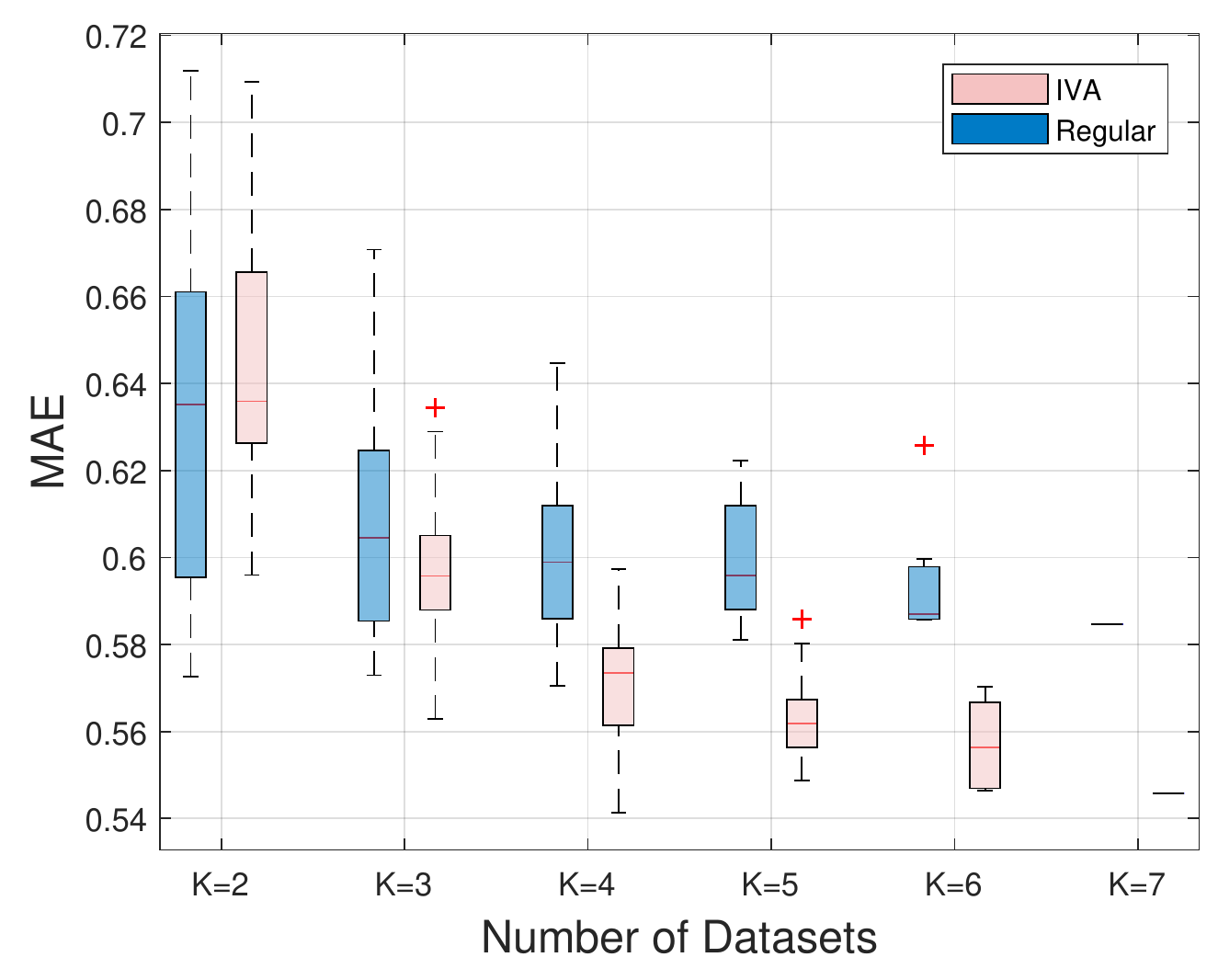}
		\caption{Boxplots of the MAE as a function of different number of datasets used in order to train the regression model. Boxplots provide statistics for the MAE when combining different featurization methods.}
		\label{BACE}
	\end{figure}

	
	Figure (\ref{BACE}) displays the MAE of the IC50 prediction as a function of the number of featurization methods used to train the regression model. We chose the dimension of the vectors created by IVA to be $P=10$ for each dataset. 
	We also excluded ICA hereafter upon observing no significant performance benefit beyond the Regular and IVA in the last section using the QM7b dataset.
	
	The MAEs in Figure (\ref{BACE}) show the improvement in the regression error with an increasing number of fused datasets.  For the sake of comparison, the errors associated with the Regular approach are also shown.  When $K=2$ the median of the Regular approach is lower than the IVA approach. However, IVA reduces the variation in the MAEs which reveals how it produces compact features. As expected, as the number of datasets is increased,  IVA performs better than the Regular approach, due to the fact that IVA exploits complementary information among the different featurization methods.
	
	In addition, Figure (\ref{BACE}) shows that as we increase the number of featurization methods, the rate the median MAE decreases is larger than the rate using the Regular approach. Finally, note that for each case the dimension of the generated feature vectors when using the IVA approach is always lower compared to the dimension of the feature vectors in the Regular approach.
	
	\section{Conclusion}
	Different featurization methods provide different information about a molecule. Thus generating features by adopting a method that exploits the inherent dependence among those different featurization methods can benefit ML tasks. In this work, we have proposed a data fusion framework that uses joint BSS techniques to exploit the underlying complementary information contained in different featurization methods. The new framework is based on IVA, a method that effectively exploits the dependence among different featurization methods and provides feature vectors that effectively can train a regression model. The IVA method that has been used is parameter free and computationally efficient with lower complexity than existing state of the art (e.g., neural network methods). In addition, its simple generative model may enable the illumination of fundamental relationships among certain characteristics of molecules, which might be of independent scientific interest. To encourage further research and application of IVA we have written the first open source Python package for IVA, pyIVA, which is available on Github.\cite{PyIVA}
	The code is based off an existing Matlab code.\cite{MatlabIVA} Currently only IVA with a multivariate Laplacian prior is implemented, but we hope our work will lead to more IVA methods being implemented in Python. 
	
	The success of the proposed method raises several interesting questions that can be explored in future work. Depending on the nature of the data we can use or develop new algorithms that take different statistical properties into account such as sparsity \cite{boukouvalas2018sparsity}. In addition, we can also compare the regression performance with other IVA algorithms or other methods such as dictionary learning \cite{mairal2009online,5714407} and non-negative matrix or tensor factorization \cite{cichocki2009nonnegative}. Lastly, order selection techniques can be used to determine the order of the reduced IVA space.
	
	\subsubsection*{Acknowledgments}
	Support for this work is gratefully acknowledged from the U.S. Office of Naval Research under grant number N00014-17-1-2108 and from the Energetics Technology Center under project number 2044-001. Partial support is also acknowledged from the Center for Engineering Concepts Development in the Department of Mechanical Engineering at the University of Maryland, College Park. We thank Dr. Ruth M. Doherty and Dr. Bill Wilson from the Energetics Technology Center for their encouragement, useful thoughts, and for proofreading the manuscript.

\bibliographystyle{unsrt}
\bibliography{references}

\end{document}